\documentclass[pdflatex,sn-mathphys-num]{sn-jnl}%

\usepackage{graphicx}%
\usepackage{multirow}%
\usepackage{amsmath,amssymb,amsfonts}%
\usepackage{amsthm}%
\usepackage{mathrsfs}%
\usepackage[title]{appendix}%
\usepackage{xcolor}%
\usepackage{textcomp}%
\usepackage{manyfoot}%
\usepackage{booktabs}%
\usepackage{graphicx}   
\usepackage{cleveref}
\usepackage[utf8]{inputenc}

\usepackage{algorithmicx}%
\usepackage{algpseudocode}%
\usepackage{listings}%
\usepackage[ruled,vlined]{algorithm2e}
\usepackage{listings}%

\usepackage{subfig}
\usepackage{tikz}
\usetikzlibrary{shapes.geometric, arrows}

\tikzstyle{startstop} = [rectangle, rounded corners, minimum width=3cm, minimum height=1cm,text centered, draw=black, fill=red!30]
\tikzstyle{process} = [rectangle, minimum width=3cm, minimum height=1cm, text centered, draw=black, fill=blue!20]
\tikzstyle{decision} = [diamond, minimum width=3cm, minimum height=1cm, text centered, draw=black, fill=green!30]
\tikzstyle{arrow} = [thick,->,>=stealth]

\UseRawInputEncoding

\begin{document}

\title{Bridging Topology and Deep Representation Learning: A TDA-ViT Fusion Model for Four-Class Brain Tumor Classification}

\author*[1]{\fnm{Faisal Ahmed} }\email{ahmedf9@erau.edu}

\affil*[1]{\orgdiv{Department of Data Science and Mathematics}, \orgname{Embry-Riddle Aeronautical University}, \orgaddress{\street{3700 Willow Creek Rd}, \city{Prescott}, \postcode{86301}, \state{Arizona}, \country{USA}}}

\abstract{
Accurate brain tumor classification from magnetic resonance imaging (MRI) is essential for supporting early diagnosis and clinical decision-making. Although Vision Transformers (ViTs) have recently achieved remarkable performance in medical image analysis by learning rich global representations, they often overlook intrinsic topological characteristics that describe the structural organization of tumor regions. To address this limitation, we propose a novel fusion framework that integrates Topological Data Analysis (TDA) features with pretrained Vision Transformer representations for four-class brain tumor MRI classification. In the proposed approach, topological descriptors extracted through TDA capture complementary geometric and structural information, while the pretrained ViT model learns high-level semantic features from MRI images. These heterogeneous feature representations are fused to form a more discriminative embedding for classification.

The proposed framework is evaluated on the BRISC2025 benchmark dataset consisting of four brain tumor categories. Experimental results demonstrate that the fusion of topological and transformer-based features significantly improves classification performance compared with individual deep learning models. The proposed TDA--ViT fusion model achieves an accuracy of 99.10\%, precision of 99.27\%, recall of 99.15\%, F1-score of 99.21\%, and an area under the receiver operating characteristic curve (AUC) of 99.98\%. Furthermore, the proposed method outperforms several state-of-the-art architectures, including ResNet50, ResNet101, EfficientNetB2, and a standalone Vision Transformer model on the same dataset. These findings demonstrate that topological information provides valuable complementary cues that enhance transformer-based feature learning, resulting in a highly robust and accurate framework for automated brain tumor classification. The proposed fusion strategy highlights the potential of combining topological representations with modern deep learning architectures for advanced medical image analysis applications.
}




\maketitle

\section{Introduction}\label{sec1}

Brain tumor classification from magnetic resonance imaging (MRI) is an important task in medical image analysis because early and accurate diagnosis can significantly improve clinical decision-making and patient outcomes. However, manual interpretation of MRI scans is time-consuming, subjective, and prone to inter-observer variability, motivating the development of automated computer-aided diagnostic systems for reliable and reproducible analysis.

In recent years, deep learning has achieved remarkable success in medical image classification tasks~\cite{litjens2017survey}. In particular, convolutional neural networks (CNNs) such as ResNet and EfficientNet have shown strong performance in brain tumor classification by learning hierarchical spatial representations from MRI data~\cite{he2016deep,tan2019efficientnet}. Despite these advances, CNN-based models are often limited in modeling long-range dependencies and global contextual relationships, which are essential for capturing the complex structural variations present in tumor regions.

To overcome these limitations, transformer-based architectures have been increasingly adopted in computer vision and medical imaging. The Vision Transformer (ViT)~\cite{dosovitskiy2021image} has demonstrated strong capability in learning global representations through self-attention mechanisms, making it especially suitable for image analysis tasks where distant spatial dependencies are important. Recent studies have also reported that transformer-based approaches can improve brain tumor classification performance, particularly when combined with hybrid feature learning or attention-based enhancements~\cite{hong2024brain, ahmed2024brain, kumar2025comprehensive}. In addition, recent reviews suggest that transformer models are becoming a major direction in MRI-based brain tumor analysis~\cite{aburass2025vision}.

Although ViT models are effective at capturing semantic and contextual information, they may still overlook intrinsic geometric and structural properties of tumor regions. Topological Data Analysis (TDA) offers a principled way to characterize shape, connectivity, and multi-scale structural patterns in medical images. By extracting topological descriptors from MRI data, TDA can provide complementary information that is not directly captured by standard deep learning features. This makes TDA particularly valuable for brain tumor classification, where subtle structural differences may distinguish between tumor types.

In this work, we propose a novel TDA--ViT fusion framework for four-class brain tumor classification using the BRISC2025 dataset. The proposed method integrates topological descriptors extracted through TDA with pretrained Vision Transformer representations to form a more discriminative feature embedding. The fusion of topological and transformer-based features aims to exploit both structural and semantic information, leading to improved classification performance and robustness.

Experimental results show that the proposed TDA--ViT model achieves 99.10\% accuracy, 99.27\% precision, 99.15\% recall, 99.21\% F1-score, and 99.98\% AUC on the BRISC2025 benchmark dataset. These results outperform several state-of-the-art deep learning models, including ResNet50, ResNet101, EfficientNetB2, and standalone ViT. The findings demonstrate that incorporating topological information into transformer-based learning provides meaningful complementary cues for automated brain tumor classification.

\noindent\textbf{Our Contributions}
\begin{itemize}
\item We propose a novel TDA--ViT fusion framework that combines topological descriptors with pretrained Vision Transformer features for brain tumor classification.
\item We evaluate the proposed model on the BRISC2025 dataset for four-class classification, including glioma, meningioma, pituitary tumors, and non-tumor cases.
\item We show that the proposed method outperforms state-of-the-art CNN-based models such as ResNet50, ResNet101, and EfficientNetB2, as well as standalone ViT.
\item We achieve 99.10\% accuracy and 99.98\% AUC, demonstrating strong discriminative capability.
\item We provide comprehensive evaluation using precision, recall, F1-score, confusion matrix, and ROC curves.
\end{itemize}

\begin{figure*}[t!]
	\centering
	
	\subfloat[\scriptsize Non-tumor MRI image.\label{fig:NT}]{
		\includegraphics[height=3.2cm,keepaspectratio]{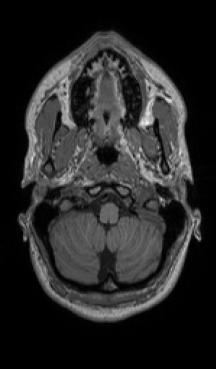}}
	\hfill
	\subfloat[\scriptsize Pituitary tumor MRI image.\label{fig:PT}]{
		\includegraphics[height=3.2cm,keepaspectratio]{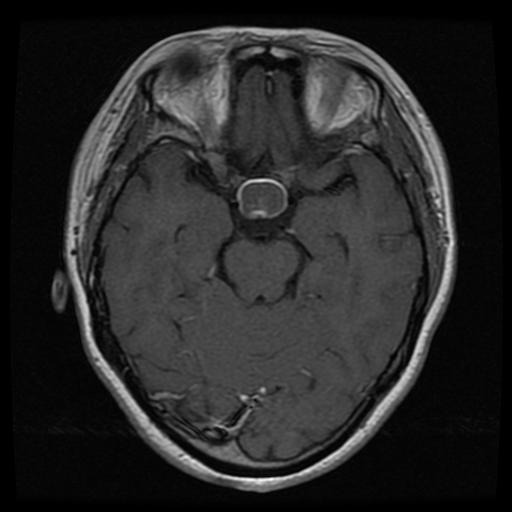}}
	\hfill
	\subfloat[\scriptsize Meningioma tumor MRI image.\label{fig:MG}]{
		\includegraphics[height=3.2cm,keepaspectratio]{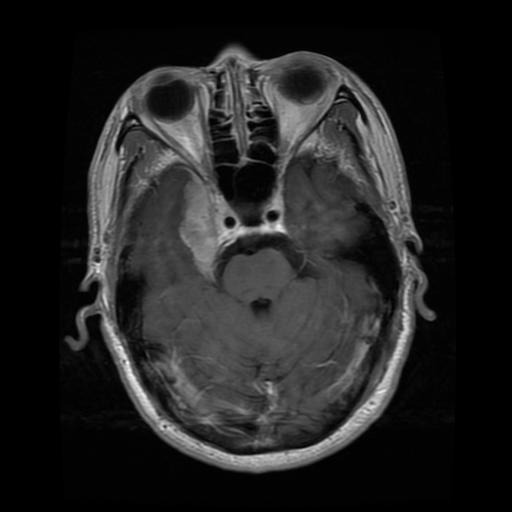}}
	\hfill
	\subfloat[\scriptsize Glioma tumor MRI image.\label{fig:GL}]{
		\includegraphics[height=3.2cm,keepaspectratio]{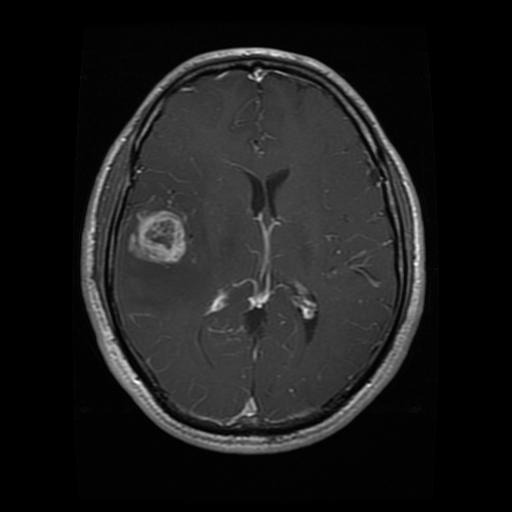}}

	\caption{\footnotesize Representative MRI samples from the BRISC2025 dataset showing the four classification categories: non-tumor, pituitary tumor, meningioma tumor, and glioma tumor. All images are resized to a uniform spatial resolution for consistent model input.}
	\label{fig:image-samples}
\end{figure*}

\section{Related Works}\label{sec2}

Automated analysis of brain magnetic resonance imaging (MRI) has become an important research direction in computer-aided diagnosis because reliable classification can support early detection, treatment planning, and clinical decision-making. In early work, brain MRI analysis was dominated by handcrafted descriptors and traditional machine learning classifiers. Structural features extracted from MRI volumes were often combined with support vector machines and related models to distinguish pathological cases from healthy controls~\cite{Kloppel2008SVM}. Although these approaches demonstrated the diagnostic value of structural MRI information, their performance was typically constrained by limited feature expressiveness and poor scalability across heterogeneous datasets.

The rise of deep learning significantly advanced brain MRI classification. Convolutional neural networks (CNNs) became the dominant paradigm because of their ability to learn hierarchical feature representations directly from imaging data~\cite{litjens2017survey}. Subsequent studies explored deeper, multi-scale, and ensemble CNN architectures to improve discrimination of local tumor patterns and contextual structures~\cite{ebrahimi2021convolutional,wang2021densecnn,fathi2024deep}. Despite these successes, CNN-based models remain biased toward local receptive fields and often fail to model long-range dependencies or global structural relationships that are important in complex brain tumor morphology. This limitation becomes particularly evident in cases where class boundaries depend on subtle structural variations rather than only local texture cues.

To improve robustness and interpretability, hybrid pipelines have also been explored. Some studies incorporated topology-preserving segmentation or feature-selection strategies to reduce redundancy and preserve anatomical consistency in MRI analysis~\cite{bazin2007topology,alshamlan2023identifying,alshamlan2024improving}. These methods improved stability and sometimes enhanced classification accuracy, yet they still relied largely on handcrafted or intensity-driven representations. As a result, they did not fully exploit the latent geometric structure embedded in MRI scans, especially when tumor morphology exhibits irregular boundaries and multi-scale spatial organization.

More recently, Topological Data Analysis (TDA) has emerged as a powerful framework for extracting robust structural information from complex biomedical data. TDA, particularly through persistent homology, characterizes connected components, holes, and higher-order geometric patterns across multiple scales~\cite{carlsson2009topology,edelsbrunner2008persistent}. Because topological descriptors are stable under small perturbations and noise, they are especially attractive for medical imaging tasks where intra-class variability and acquisition differences are common. In brain MRI applications, TDA has shown promise in capturing structural signatures that are complementary to intensity-based or convolutional features. Several studies have shown that Topological Data Analysis (TDA) can effectively capture robust structural and shape-based information in medical imaging ~\cite{ahmed2026four,ahmed2025topo,ahmed2023topo,ahmed2023topological,ahmed2023tofi,ahmed20253d,yadav2023histopathological,ahmed2025topological,ahmed2026brain,ahmed2026hybrid}. This makes TDA a compelling choice for tumor classification tasks in which morphological organization carries discriminative value.

In parallel, transformer-based architectures have reshaped computer vision by enabling global representation learning through self-attention. The Vision Transformer (ViT) has demonstrated that patch-based tokenization and attention can achieve strong performance in image recognition tasks~\cite{dosovitskiy2021image}. Variants such as Swin Transformer further improved efficiency and hierarchical feature learning by introducing shifted window attention~\cite{liu2021swin}. In medical imaging, transformer-based models have increasingly been adopted for MRI classification because they can model long-range dependencies more effectively than CNNs~\cite{sankari2025hierarchical,dhinagar2023efficiently}. However, ViTs typically require substantial training data and may not fully capture domain-specific structural cues when used alone, particularly in grayscale medical images where subtle topological differences are diagnostically meaningful. To address these challenges, recent studies have introduced hybrid and enhanced transformer-based frameworks that improve feature representation, robustness, and generalization in medical image classification~\cite{ahmed2025colormap,ahmed2026hog,ahmed2025ocuvit,ahmed2025robust,ahmed2025histovit,ahmed2025addressing,ahmed2025repvit,ahmed2025pseudocolorvit,rawat2025efficient,ahmed2025transfer}.

Recent literature has therefore shifted toward hybrid and enhanced transformer-based methods that attempt to improve representation quality and generalization in medical image analysis. These studies suggest that combining complementary feature sources can lead to more robust performance than using a single backbone alone. In this context, the integration of TDA with ViT is especially promising because it merges two fundamentally different but complementary perspectives: TDA captures structural and topological organization, while ViT learns semantic and global contextual representations. Such a fusion is well aligned with the needs of brain tumor MRI classification, where the discriminative signal may arise from both morphology and higher-level appearance cues.

Motivated by these observations, the proposed TDA--ViT framework fuses topological descriptors with pretrained transformer features to produce a richer and more discriminative embedding for four-class brain tumor classification. Compared with CNN-only and transformer-only approaches, this design aims to improve robustness, reduce the risk of overlooking structural cues, and better exploit the complex geometry of tumor regions in MRI scans. The literature reviewed above supports the premise that topology-aware transformer fusion is a promising direction for advancing reliable and accurate medical image classification.

\begin{figure*}[t!]
    \centering
    \includegraphics[width=\linewidth]{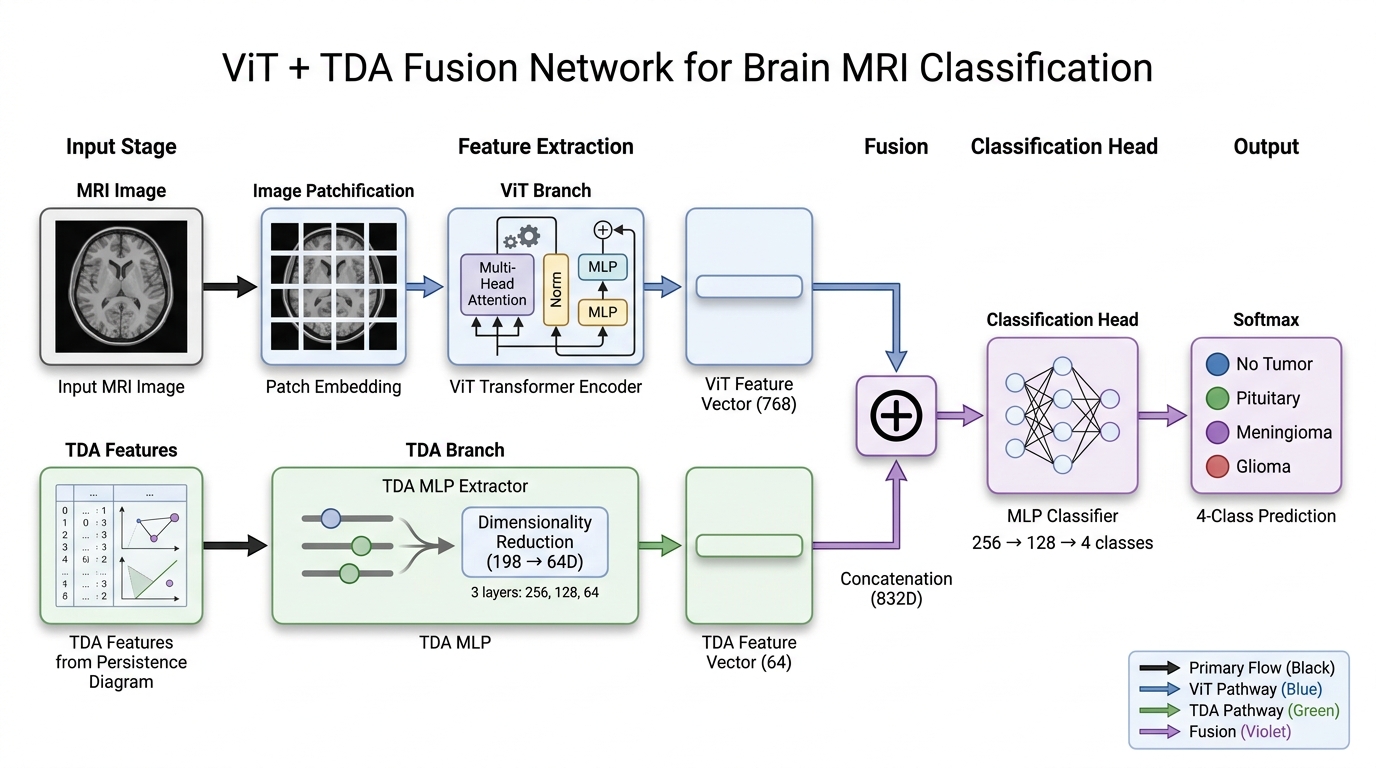}
    \caption{\footnotesize
    \textbf{Architecture of the proposed TDA--ViT fusion framework for four-class brain tumor MRI classification.}
    The framework consists of two complementary feature extraction branches. In the Vision Transformer (ViT) branch, each MRI image is resized and partitioned into non-overlapping patches, which are embedded and processed through multiple transformer encoder layers to produce a 768-dimensional deep semantic feature vector. In parallel, the Topological Data Analysis (TDA) branch extracts persistent homology-based descriptors from MRI images, generating a 198-dimensional topological feature vector that is subsequently refined through a multilayer perceptron (MLP) to obtain a compact 64-dimensional representation. The features from both branches are concatenated to form an 832-dimensional fused feature vector, which is passed through a fully connected classification head for final prediction. The Softmax output layer produces probability scores for the four diagnostic categories: non-tumor, pituitary tumor, meningioma, and glioma.}
    \label{fig:flowchart}
\end{figure*}
\section{Methodology}\label{sec:method}

This section presents the proposed TDA--ViT fusion framework for four-class brain tumor MRI classification using the BRISC2025 dataset (representative sample in \Cref{fig:image-samples}). The proposed model integrates topological descriptors extracted through persistent homology with deep semantic representations learned by a pretrained Vision Transformer (ViT). The overall pipeline consists of three major stages: (i) Vision Transformer feature extraction, (ii) Topological Data Analysis (TDA) feature extraction, and (iii) feature fusion and classification. The complete workflow of the proposed framework is illustrated in \Cref{fig:flowchart}.

\subsection{Vision Transformer Feature Extraction}

Let $\mathbf{I}\in\mathbb{R}^{H\times W}$ denote an input MRI image. Each image is first resized to a fixed resolution of $224\times224$ pixels to satisfy the input requirement of the pretrained Vision Transformer model.

Pixel intensities are normalized as

\[
\mathbf{I}_{norm}
=
\frac{\mathbf{I}}{255},
\qquad
\mathbf{I}_{norm}
\in
[0,1]^{224\times224}.
\]

Since the pretrained ViT model expects three-channel images, the normalized grayscale MRI image is replicated across three channels:

\[
\mathbf{X}
=
[\mathbf{I}_{norm},
\mathbf{I}_{norm},
\mathbf{I}_{norm}]
\in
\mathbb{R}^{224\times224\times3}.
\]

The image is then divided into non-overlapping patches of size $16\times16$. For an image of size $224\times224$, the number of patches is

\[
N
=
\frac{224\times224}
{16\times16}
=
196.
\]

Thus,

\[
\mathbf{X}
\rightarrow
\{
\mathbf{x}_1,
\mathbf{x}_2,
\dots,
\mathbf{x}_N
\}.
\]

Each patch is flattened and projected into a latent embedding space:

\[
\mathbf{e}_i
=
\mathbf{W}_e
\,
\text{vec}
(\mathbf{x}_i)
+
\mathbf{b}_e.
\]

A learnable classification token $\mathbf{x}_{cls}$ is prepended to the patch sequence and positional embeddings are added:

\[
\mathbf{z}^{0}
=
[
\mathbf{x}_{cls},
\mathbf{e}_1+\mathbf{p}_1,
\dots,
\mathbf{e}_N+\mathbf{p}_N
].
\]

The sequence is processed through $L$ transformer encoder layers. Each layer consists of Multi-Head Self-Attention (MSA) and a Feed-Forward Network (FFN).

The self-attention operation is defined as

\[
\text{MSA}(\mathbf{Z})
=
\text{Softmax}
\left(
\frac{\mathbf{QK}^{T}}
{\sqrt{d}}
\right)
\mathbf{V},
\]

where $\mathbf{Q}$, $\mathbf{K}$ and $\mathbf{V}$ denote the query, key and value matrices.

The transformer encoder updates the latent representation as

\[
\mathbf{Z}'
=
\mathbf{Z}
+
\text{MSA}(\mathbf{Z}),
\]

\[
\mathbf{Z}^{l+1}
=
\mathbf{Z}'
+
\text{FFN}
(\mathbf{Z}').
\]

After the final encoder layer, the representation corresponding to the classification token is extracted:

\[
\mathbf{f}_{ViT}
=
\mathbf{z}^{L}_{cls},
\]

where

\[
\mathbf{f}_{ViT}
\in
\mathbb{R}^{768}.
\]

This vector serves as the deep semantic representation of the MRI image and is subsequently used as one component of the proposed fusion framework. When evaluated independently using the same training and testing protocol, the pretrained ViT model achieved an accuracy of 98.90\%, demonstrating its strong capability for four-class brain tumor classification. 

\subsection{Topological Data Analysis Feature Extraction}

To capture structural characteristics of brain tumors, persistent homology is employed as a topological feature extraction technique.

For a grayscale MRI image

\[
\mathcal{X}
\in
\mathbb{R}^{r\times s},
\]

a cubical filtration is constructed using pixel intensity values.

The filtration generates a sequence of nested cubical complexes:

\[
\mathcal{X}_1
\subset
\mathcal{X}_2
\subset
\dots
\subset
\mathcal{X}_N,
\]

where

\[
\mathcal{X}_n
=
\{
\Delta_{ij}
\subset
\mathcal{X}
\mid
\gamma_{ij}
\le
t_n
\}.
\]

Here,
$t_1<t_2<\dots<t_N$
represent filtration thresholds spanning the grayscale intensity range.

\begin{figure}[t]
    \centering
    \includegraphics[width=\linewidth]{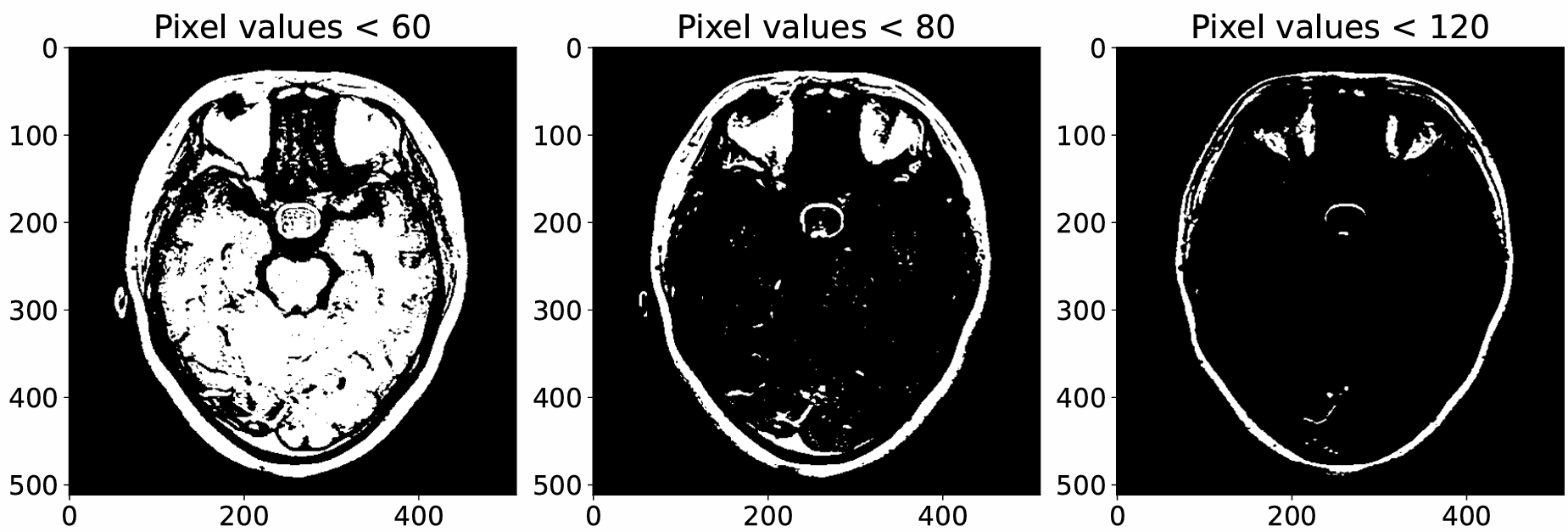}
    \caption{\small 
    \textbf{Sublevel filtration of an MRI image.} 
   Binary images $\mathcal{X}_{60}$, $\mathcal{X}_{80}$, and $\mathcal{X}_{120}$ obtained at filtration thresholds 60, 80, and 120, illustrating the progressive evolution of topological structures.
    }
    \label{fig:filtration}
\end{figure}

For each filtration (see \Cref{fig:filtration}), persistence diagrams are computed to track the birth and death of topological structures.

The persistence diagram for homology dimension $k$ is

\[
PD_k(\mathcal{X})
=
\{
(b_{\sigma},d_{\sigma})
\mid
\sigma
\in
H_k
\},
\]

where $b_{\sigma}$ and $d_{\sigma}$ denote the birth and death times of a topological feature.

In this study, both

\[
H_0
\quad
\text{and}
\quad
H_1
\]

are computed.

The resulting persistence diagrams are transformed into Betti curves:

\[
\beta_k(t)
=
\sum_{i=1}^{m}
\mathbb{I}
(
b_i
\le
t
<
d_i
).
\]

Betti curves provide a compact summary of the number of active topological structures throughout the filtration.

For computational efficiency, the Betti curves are discretized into fixed-length vectors.

The Betti-0 and Betti-1 vectors are concatenated to obtain

\[
\mathbf{f}_{TDA}
\in
\mathbb{R}^{198}.
\]

The extracted TDA features are then processed using a dedicated neural network branch consisting of fully connected layers:

\[
\mathbf{h}_1
=
\text{ReLU}
(
\mathbf{W}_1
\mathbf{f}_{TDA}
+
\mathbf{b}_1
),
\]

\[
\mathbf{h}_2
=
\text{ReLU}
(
\mathbf{W}_2
\mathbf{h}_1
+
\mathbf{b}_2
),
\]

yielding

\[
\mathbf{h}_{TDA}
\in
\mathbb{R}^{64}.
\]

To evaluate the discriminative capability of the extracted topological descriptors independently of deep learning, the concatenated Betti-0 and Betti-1 feature vectors were also used to train an XGBoost classifier. Using only the handcrafted topological features, the model achieved an accuracy of 98.19\% on the BRISC2025 dataset, demonstrating that persistent homology captures highly informative structural characteristics of brain tumors. Nevertheless, the proposed TDA branch learns a more compact and task-specific representation that can be effectively integrated with transformer-based features for improved classification performance.
\subsection{Feature Fusion and Classification}

The deep semantic representation obtained from the Vision Transformer and the topological representation obtained from the TDA branch are fused through feature concatenation.

Let

\[
\mathbf{f}_{ViT}
\in
\mathbb{R}^{768}
\]

and

\[
\mathbf{h}_{TDA}
\in
\mathbb{R}^{64}.
\]

The fused representation is

\[
\mathbf{f}_{fusion}
=
[
\mathbf{f}_{ViT};
\mathbf{h}_{TDA}
]
\in
\mathbb{R}^{832}.
\]

The fused feature vector is passed through a multilayer classifier:

\[
\mathbf{z}_1
=
\text{ReLU}
(
\mathbf{W}_3
\mathbf{f}_{fusion}
+
\mathbf{b}_3
),
\]

\[
\mathbf{z}_2
=
\text{ReLU}
(
\mathbf{W}_4
\mathbf{z}_1
+
\mathbf{b}_4
).
\]

Finally,

\[
\hat{\mathbf{y}}
=
\text{Softmax}
(
\mathbf{W}_5
\mathbf{z}_2
+
\mathbf{b}_5
)
\]

produces probabilities for the four brain tumor classes.

The network is optimized using the categorical cross-entropy loss:

\[
\mathcal{L}
=
-\frac{1}{B}
\sum_{i=1}^{B}
\sum_{c=1}^{4}
y_{i,c}
\log
(
\hat{y}_{i,c}
).
\]

\begin{figure*}[t!]
    \centering
    \subfloat[\scriptsize One-vs-rest ROC curves with corresponding AUC values for the four brain tumor classes in the BRISC2025 dataset.\label{fig:auc}]{
        \includegraphics[width=0.45\linewidth]{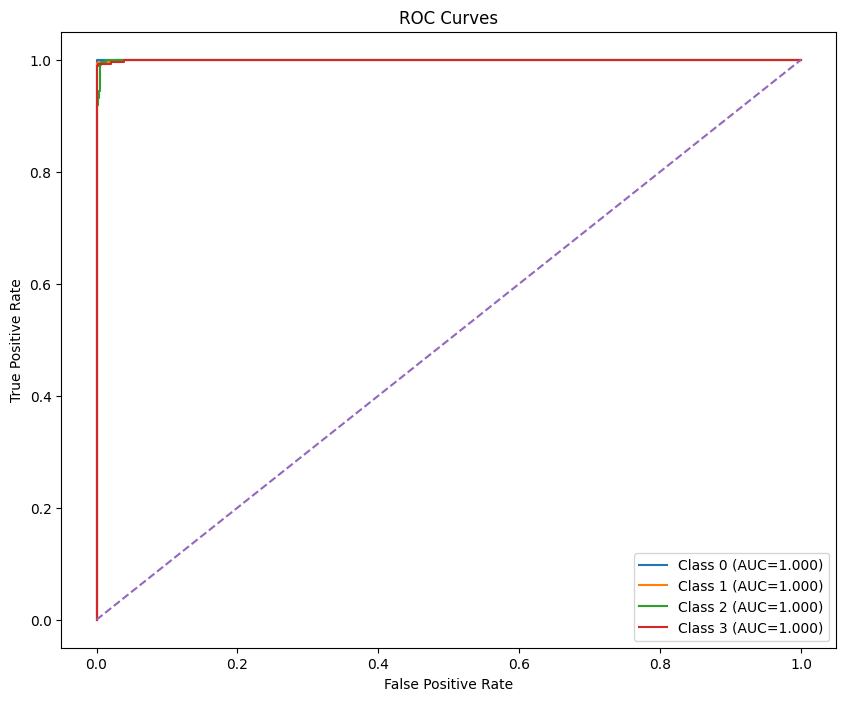}}
    \hfill
    \subfloat[\scriptsize Confusion matrix showing class-wise classification performance of the proposed TDA--ViT fusion model.\label{fig:conf}]{
        \includegraphics[width=0.45\linewidth]{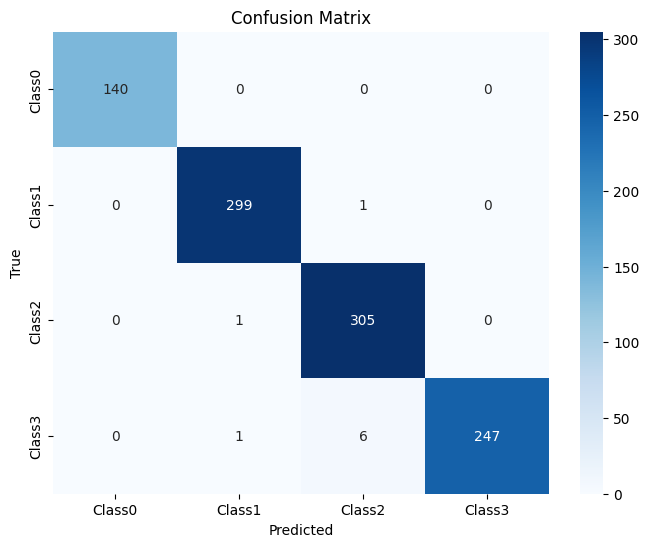}}
    
    \caption{\footnotesize
    Performance evaluation of the proposed \textbf{TDA--ViT fusion framework} on the BRISC2025 brain tumor MRI dataset. (a) One-vs-rest receiver operating characteristic (ROC) curves demonstrating the discriminative capability of the model across the glioma, meningioma, pituitary, and non-tumorous classes, achieving near-perfect AUC scores. (b) Confusion matrix illustrating the distribution of correct and incorrect predictions, highlighting the strong classification performance and minimal inter-class confusion achieved by the proposed approach.}
    
    \label{fig:auc_conf}
\end{figure*}

The fusion of topological descriptors and transformer-based deep representations enables the proposed framework to simultaneously capture structural, geometric, and semantic characteristics of brain tumors. By leveraging the complementary strengths of TDA and Vision Transformers, the model achieves enhanced discriminative capability and superior classification performance on the BRISC2025 dataset. The algorithm of this model is given in Alg.\ref{alg:tdavit}. The resulting classification effectiveness is further illustrated through the confusion matrix and one-vs-rest ROC analysis presented in \Cref{fig:auc_conf}.

\subsection{Hyperparameter Settings}

The proposed TDA--ViT fusion framework involves several hyperparameters associated with image preprocessing, topological feature learning, transformer-based feature extraction, and classification. These parameters were selected empirically to achieve stable training and robust classification performance on the BRISC2025 dataset. Table~\ref{tab:hyperparameters} summarizes the hyperparameter configuration used throughout all experiments.

\begin{table}[h!]
\centering
\caption{Hyperparameter configuration of the proposed TDA--ViT fusion framework.}
\label{tab:hyperparameters}
\setlength{\tabcolsep}{6pt}
\begin{tabular}{ll}
\toprule
\textbf{Hyperparameter} & \textbf{Value} \\
\midrule
ViT Backbone & ViT-Base-Patch16-224 \\
Input Image Size & $224 \times 224$ \\
Patch Size & $16 \times 16$ \\
Number of Patches & 196 \\
Number of Classes & 4 \\
Batch Size & 16 \\
Maximum Epochs & 50 \\
Optimizer & AdamW \\
Learning Rate & $1 \times 10^{-4}$ \\
Weight Decay & $1 \times 10^{-4}$ \\
Loss Function & Cross-Entropy Loss \\
Early Stopping Patience & 10 \\
TDA Feature Dimension & 198 \\
TDA Embedding Dimension & 64 \\
ViT Feature Dimension & 768 \\
Fusion Feature Dimension & 832 \\
Dropout (TDA Branch) & 0.3 \\
Dropout (Fusion Classifier) & 0.4, 0.3 \\
Weight Initialization & Pretrained ImageNet Weights \\
Training Shuffle & True \\
Computing Device & CPU/GPU (Auto-detected) \\
\bottomrule
\end{tabular}
\end{table}

\begin{algorithm}[t]
\SetAlgoNlRelativeSize{0}
\DontPrintSemicolon
\caption{TDA--ViT Fusion Framework for Four-Class Brain Tumor MRI Classification}
\label{alg:tdavit}

\KwIn{MRI dataset $\mathcal{D}=\{(\mathbf{I}_i,y_i)\}_{i=1}^{N}$, pretrained ViT, maximum epochs $E$, patience $p$, number of classes $C=4$}
\KwOut{Trained TDA--ViT model and classification metrics}

\textbf{MRI Preprocessing:}

\For{$i=1$ \KwTo $N$}{
Convert MRI image to grayscale;

Resize image to $224\times224$;

Normalize pixel intensities to $[0,1]$;

Replicate grayscale image into three channels;

Generate tensor representation
$\mathbf{X}_i\in\mathbb{R}^{3\times224\times224}$;
}

\vspace{0.1cm}

\textbf{Topological Feature Extraction:}

\For{$i=1$ \KwTo $N$}{

Construct cubical filtration from MRI image $\mathbf{I}_i$;

Compute persistence diagrams for $H_0$ and $H_1$;

Generate Betti-0 and Betti-1 curves;

Vectorize and concatenate Betti features;

Obtain TDA feature vector

$\mathbf{t}_i\in\mathbb{R}^{198}$;
}

\vspace{0.1cm}

\textbf{Model Initialization:}

Load pretrained ViT-Base-Patch16-224;

Initialize TDA branch network

$(198 \rightarrow 256 \rightarrow 128 \rightarrow 64)$;

Initialize fusion classifier

$(832 \rightarrow 256 \rightarrow 128 \rightarrow C)$;

Initialize AdamW optimizer

$(\eta=10^{-4},\ \lambda=10^{-4})$;

Define cross-entropy loss function;

\vspace{0.1cm}

\textbf{Training:}

Initialize best accuracy $\alpha^{*}=0$;

Initialize early stopping counter $\delta=0$;

\For{epoch $=1$ \KwTo $E$}{

\ForEach{mini-batch $(\mathbf{X},\mathbf{t},y)$}{

Extract ViT features

$\mathbf{f}_{ViT}\in\mathbb{R}^{768}$;

Extract topological embedding

$\mathbf{h}_{TDA}\in\mathbb{R}^{64}$;

Fuse representations

$\mathbf{f}_{fusion}
=
[\mathbf{f}_{ViT};\mathbf{h}_{TDA}]$;

Compute class probabilities using Softmax;

Calculate cross-entropy loss;

Perform backpropagation;

Update network parameters using AdamW;
}

Evaluate model on the test set;

\If{Accuracy $>\alpha^{*}$}{

Save model parameters;

Update $\alpha^{*}$;

Reset $\delta=0$;
}

\Else{

$\delta \gets \delta + 1$;
}

\If{$\delta \ge p$}{

Terminate training;
}
}

\vspace{0.1cm}

\textbf{Evaluation:}

Load best-performing model;

Compute prediction probabilities;

Generate final class labels;

Calculate Accuracy, Precision, Recall, F1-score, and AUC;

Construct confusion matrix;

Generate one-vs-rest ROC curves;

\Return Best TDA--ViT model and evaluation metrics;

\end{algorithm}

\section{Experiment}

\subsection{Dataset}

\noindent \textbf{BRISC2025 Dataset}~\cite{brisc2025}

The BRISC2025 dataset is a publicly available benchmark developed for brain tumor classification and segmentation using magnetic resonance imaging (MRI). The dataset comprises 6,000 contrast-enhanced T1-weighted MRI images annotated by experienced radiologists and medical experts. It contains four diagnostic categories: glioma, meningioma, pituitary tumor, and non-tumorous brain images. In addition to class labels, each MRI scan is accompanied by pixel-level segmentation masks, facilitating both classification and tumor localization studies.

The images are collected from multiple anatomical perspectives, including axial, sagittal, and coronal planes, thereby providing substantial variability in tumor appearance and spatial characteristics. This diversity supports the development of robust deep learning models capable of generalizing across different imaging views.

For experimental evaluation, the dataset is partitioned into training and testing subsets containing 5,000 and 1,000 images, respectively. Table~\ref{tab:brisc_distribution} summarizes the class-wise distribution of samples in the training and testing sets. The relatively balanced distribution among the four categories makes BRISC2025 a suitable benchmark for evaluating automated brain tumor classification systems.

\begin{table}[ht]
\centering
\caption{Class distribution of the BRISC2025 dataset.}
\label{tab:brisc_distribution}
\setlength{\tabcolsep}{8pt}
\begin{tabular}{lccc}
\toprule
\textbf{Class} & \textbf{Training Images} & \textbf{Testing Images} & \textbf{Total Images} \\
\midrule
Glioma        & 1,147 & 254 & 1,401 \\
Meningioma    & 1,329 & 306 & 1,635 \\
Pituitary     & 1,457 & 300 & 1,757 \\
Non-tumorous  & 1,067 & 140 & 1,207 \\
\midrule
\textbf{Total} & \textbf{5,000} & \textbf{1,000} & \textbf{6,000} \\
\bottomrule
\end{tabular}
\end{table}

\subsection{Experimental Setup}

\noindent \textbf{Training--Test Split:}
To ensure a fair and reproducible evaluation, we adopted the predefined training and testing partitions provided by the BRISC2025 dataset. As summarized in Table~\ref{tab:brisc_distribution}, the dataset contains 5,000 training images and 1,000 testing images distributed across four classes: glioma, meningioma, pituitary tumor, and non-tumorous cases. Utilizing the official dataset split enables direct comparison with previously reported results while maintaining consistency with established benchmarking protocols for four-class brain tumor MRI classification. All model training, validation, and performance evaluation procedures were conducted using these standardized data partitions.

\smallskip

\noindent \textbf{No Data Augmentation:}
In contrast to many deep learning studies that employ extensive data augmentation strategies to improve model robustness and mitigate overfitting~\cite{goutam2022comprehensive}, the proposed TDA--ViT fusion framework was trained using only the original MRI images provided in the BRISC2025 dataset. No artificial transformations, such as image rotation, flipping, scaling, translation, or intensity perturbation, were applied during training.



\noindent \textbf{Computational Platform:}
All experiments were performed on a personal laptop equipped with an Apple M1 system-on-chip (SoC), comprising an 8-core CPU with four performance cores and four efficiency cores, and 16~GB of unified memory. The proposed TDA--ViT framework was trained and evaluated using this resource-constrained computing environment, demonstrating its practical applicability without requiring access to high-end GPU infrastructure.


\section{Results}\label{sec:results}

The performance of the proposed TDA--ViT fusion framework is evaluated on the BRISC2025 dataset for four-class brain tumor MRI classification. The model is assessed using standard evaluation metrics, including precision, recall, accuracy, F1-score, and area under the receiver operating characteristic curve (AUC). A comparative analysis with state-of-the-art deep learning models is also presented to demonstrate the effectiveness of the proposed approach.

Table~\ref{tab:results} summarizes the quantitative results. Among the baseline models, ResNet50 achieves an accuracy of 98.20\%, while ResNet101 records 98.09\%. EfficientNetB2 slightly improves the performance with an accuracy of 98.37\%. A CNN model trained on the Kaggle dataset reports an accuracy of 97.24\%, while the standalone Vision Transformer (ViT) achieves a stronger accuracy of 98.90\% with an AUC of 99.97\%. Although these methods perform well, the proposed TDA--ViT framework attains the best results across all metrics.

Specifically, the proposed method achieves a precision of 99.27\%, recall of 99.15\%, accuracy of 99.10\%, F1-score of 99.21\%, and AUC of 99.98\%. This consistent improvement indicates that integrating topological descriptors with transformer-based representations produces a more discriminative feature space for brain tumor classification. The gain over the standalone ViT model also confirms that TDA contributes complementary structural information beyond global semantic features.

The superior performance of the proposed framework can be attributed to two main factors. First, the Vision Transformer captures long-range contextual relationships within MRI images through its self-attention mechanism, overcoming the locality limitations of conventional CNNs. Second, the TDA component encodes intrinsic geometric and topological properties of tumor regions, which helps preserve structural information that may otherwise be overlooked by purely deep feature extractors. The fusion of these two representations leads to improved separability among the four tumor classes.

Further analysis using the confusion matrix shows that the model achieves highly accurate classification across all classes, with only a small number of misclassified samples. The ROC curves also demonstrate excellent class-wise separability, as reflected by the near-perfect AUC values in the one-vs-rest setting. These results indicate that the proposed TDA--ViT framework is both robust and reliable for multi-class brain tumor MRI classification.

Overall, the experimental results show that the proposed approach outperforms traditional CNN-based models as well as the standalone ViT model. This confirms that combining topological information with transformer-based learning is a promising strategy for advanced medical image classification tasks.

\begin{table*}[h!]
\centering
\caption{\footnotesize 
Comparison of the proposed TDA--ViT fusion framework with state-of-the-art deep learning models for four-class brain tumor MRI classification on the BRISC2025 dataset. Reported metrics include macro-averaged precision (Prec), recall (Rec), accuracy, F1-score, and one-vs-rest area under the receiver operating characteristic curve (AUC). The proposed method achieves the highest performance across all evaluation metrics, demonstrating the effectiveness of integrating topological and transformer-based representations for brain tumor classification.
\label{tab:results}}
\setlength\tabcolsep{4pt}
\footnotesize

\begin{tabular}{lccccccc}
\multicolumn{8}{c}{\bf BRISC2025 Dataset: Four-Class Brain Tumor MRI Classification Results} \\
\toprule
Method & \# Classes & Dataset & Prec & Rec & Accuracy & F1-score & AUC \\
\midrule

ResNet50~\cite{fateh2026brisc} 
& 4 & BRISC2025 & 98.36 & 98.36 & 98.20 & 98.34 & - \\

ResNet101~\cite{fateh2026brisc} 
& 4 & BRISC2025 & 98.20 & 98.26 & 98.09 & 98.21 & - \\

EfficientNetB2~\cite{fateh2026brisc} 
& 4 & BRISC2025 & 98.51 & 98.54 & 98.37 & 98.52 & - \\
CNN ~\cite{indian2026improved}
& 4 & Kaggle & 97.26 & 97.24 & 97.24 & 97.24 & 99.42 \\
ViT ~\cite{ahmed2026enhancing}
& 4 & BRISC2025 & 98.99 & 98.98 & 98.90 & 98.98 & 99.97 \\

\midrule

\bf TDA+ViT
& 4 & BRISC2025 & \textbf{99.27} & \textbf{99.15} & \textbf{99.10} & \textbf{99.21} & \textbf{99.98} \\

\bottomrule
\end{tabular}
\end{table*}

\section{Discussion}\label{sec:discussion}

The experimental results show that the proposed TDA--ViT fusion framework consistently outperforms conventional CNN-based architectures as well as the standalone Vision Transformer. The improvements observed in accuracy, precision, recall, and F1-score indicate that the proposed model learns a more discriminative representation of brain MRI data by combining complementary structural and semantic information.

A major advantage of the proposed approach is the ability of the Vision Transformer to capture long-range dependencies and global contextual relationships through self-attention. Unlike CNNs, which are generally constrained by local receptive fields, ViT models can better represent spatial interactions across distant regions of an image. This capability is particularly important for brain tumor MRI analysis, where class-relevant patterns may be distributed across multiple anatomical areas rather than confined to a single local region.

The inclusion of TDA features further strengthens the model by encoding intrinsic geometric and topological properties of tumor regions. These descriptors capture structural information such as connectedness, shape variation, and multi-scale organization, which are often difficult for purely appearance-based deep learning models to exploit. As a result, the fusion of TDA with ViT produces a richer embedding that improves class separability and enhances classification robustness.

The high AUC score obtained by the proposed method confirms that the model achieves strong discriminative performance across all classes. The confusion matrix also shows only limited misclassification, suggesting that the proposed framework generalizes well within the BRISC2025 dataset. In addition, the use of pretrained ViT representations helps stabilize training and reduce optimization difficulty, while the TDA branch contributes domain-specific structural cues that are not explicitly modeled by the transformer alone.

Despite these promising results, some limitations remain. The proposed framework depends on pretrained transformer features, which may increase computational cost and memory requirements. In addition, the evaluation is currently restricted to the BRISC2025 dataset, so further validation on independent external datasets is necessary to confirm generalizability under different imaging conditions and clinical settings. Future work may also explore end-to-end fusion strategies, lightweight transformer backbones, and multimodal extensions to further improve efficiency and robustness.

Overall, the findings suggest that combining topological information with transformer-based representation learning is an effective strategy for brain tumor MRI classification. The proposed TDA--ViT framework demonstrates that structural and semantic feature fusion can provide a more reliable and accurate solution for medical image analysis.

\section{Conclusion}\label{sec:conclusion}

In this study, we proposed a novel TDA--ViT fusion framework for multi-class brain tumor classification from MRI data. The proposed method combines the global representation learning capability of Vision Transformers with topological descriptors extracted through TDA, enabling the model to capture both semantic and structural information from brain MRI scans. This complementary fusion strategy improves the discriminative power of the learned embedding and enhances classification performance. Experimental evaluation on the BRISC2025 dataset demonstrates that the proposed framework outperforms several state-of-the-art deep learning models, including ResNet50, ResNet101, EfficientNetB2, and the standalone Vision Transformer. The proposed TDA--ViT model achieves an accuracy of 99.10\%, precision of 99.27\%, recall of 99.15\%, F1-score of 99.21\%, and AUC of 99.98\%. These results confirm that topological information provides valuable complementary cues that improve robustness, reliability, and overall classification accuracy.

The findings of this work highlight the potential of combining topology-aware feature extraction with transformer-based representation learning for medical image analysis. In particular, the proposed approach demonstrates that structural geometry and global contextual modeling can be jointly exploited to address the challenges of brain tumor classification more effectively than conventional CNN-based methods.

For future work, the framework can be extended to multimodal MRI data, lightweight transformer architectures, and larger external datasets to further assess generalizability. Additional directions include end-to-end trainable fusion strategies, explainability analysis, and validation in broader clinical settings to support real-world deployment in decision-support systems.

\clearpage

 \section*{Declarations}

 \textbf{Funding} \\
 The author received no financial support for the research, authorship, or publication of this work.

 \vspace{2mm}
 \textbf{Author's Contribution} \\
 Faisal Ahmed conceptualized the study, downloaded the data, prepared the code, performed the data analysis and wrote the manuscript. Faisal Ahmed reviewed and approved the final version of the manuscript. 

  \vspace{2mm}
 \textbf{Acknowledgement} \\
The authors utilized an online platform to check and correct grammatical errors and to improve sentence readability.

 \vspace{2mm}
 \textbf{Conflict of interest/Competing interests} \\
 The authors declare no conflict of interest.

 \vspace{2mm}
 \textbf{Ethics approval and consent to participate} \\
 Not applicable. This study did not involve human participants or animals, and publicly available datasets were used.

 \vspace{2mm}
 \textbf{Consent for publication} \\
 Not applicable.

 \vspace{2mm}
 \textbf{Data availability} \\
 The datasets used in this study are publicly available online. 

 \vspace{2mm}
 \textbf{Materials availability} \\
 Not applicable.


\bibliography{refs}
\end{document}